\documentclass[11pt,a4paper]{article}
\usepackage[hyperref]{acl2019}
\usepackage{times}
\usepackage{latexsym}

\usepackage{url}

\usepackage{multirow}
\usepackage{array}
\usepackage{xcolor}
\usepackage{ulem}
\usepackage{graphicx}
\usepackage{amsmath}
\usepackage{amssymb}
\usepackage{makecell}
\usepackage{algorithm}
\usepackage{algpseudocode}

\aclfinalcopy 

\title{
A Neural Grammatical Error Correction System Built On \\
Better Pre-training and Sequential Transfer Learning 
}

\author{
  Yo Joong Choe\thanks{\;\;Equal contribution. Listed alphabetically.} \\
  Kakao \\
  \texttt{yj.c@kakaocorp.com} \\\And
  Jiyeon Ham\textsuperscript{*} \\
  Kakao Brain \\
  \texttt{jiyeon.ham@kakaobrain.com} \\\AND
  Kyubyong Park\textsuperscript{*} \\
  Kakao Brain \\
  \texttt{kyubyong.park@kakaobrain.com} \\\And
  Yeoil Yoon\textsuperscript{*} \\
  Kakao Brain \\
  \texttt{yeoil.yoon@kakaobrain.com} \\
  }

\date{}

\begin{document}
\maketitle
\begin{abstract}
Grammatical error correction can be viewed as a low-resource sequence-to-sequence task, because publicly available parallel corpora are limited. 
To tackle this challenge, we first generate erroneous versions of large unannotated corpora using a realistic noising function.
The resulting parallel corpora are subsequently used to pre-train Transformer models.
Then, by sequentially applying transfer learning, we adapt these models to the domain and style of the test set. 
Combined with a context-aware neural spellchecker, our system achieves competitive results in both restricted and low resource tracks in ACL 2019 BEA Shared Task. 
We release all of our code and materials for reproducibility. \footnote{\url{https://github.com/kakaobrain/helo\_word}}
\end{abstract}

\section{Introduction}

Grammatical error correction (GEC) is the task of correcting various grammatical errors in text, as illustrated by the following example:
\begin{center}
    [Travel $\rightarrow$ \textbf{Travelling}] by bus is [exspensive $\rightarrow$ \textbf{expensive}], [bored $\rightarrow$ \textbf{boring}] and annoying.
\end{center}

While the dominant approach following the CoNLL-2014 Shared Task \cite{ng2014conll} has been different adaptations of phrase-based and statistical machine translation (PBSMT) models \cite{junczys2016phrase}, more recent work on GEC increasingly adopted partial \cite{grundkiewicz2018near} or exclusive \cite{junczys2018approaching, chollampatt2018multilayer} use of deep sequence-to-sequence (seq2seq) architectures \cite{sutskever2014sequence, cho2014learning}, which showed immense success in neural machine translation (NMT) \cite{bahdanau2015neural, gehring2017convolutional, vaswani2017attention}.

In GEC, unlike NMT between major languages, there are not enough publicly available corpora (GEC's hundreds of thousands to NMT's tens of millions). 
This motivates the use of pre-training and transfer learning, which has shown to be highly effective in many natural language processing (NLP) scenarios in which there is not enough annotated data, notably in low-resource machine translation (MT) \cite{lample2018phrase, ruder2019neural}.
As a result, recent GEC systems also include pre-training on various auxiliary tasks, such as language modeling (LM) \cite{junczys2018approaching}, text revision \cite{lichtarge2018weakly}, and denoising \cite{zhao2019improving}.

In this paper, we introduce a neural GEC system that combines the power of pre-training and transfer learning.
Our contributions are summarized as follows:
\begin{itemize}
    \item We pre-train our model for the denoising task using a novel noising function, which gives us a parallel corpus that includes realistic grammatical errors;
    \item We leverage the idea of sequential transfer learning \cite{ruder2019neural}, thereby effectively adapting our pre-trained model to the domain as well as the writing and annotation styles suitable for our final task.
    \item We introduce a context-aware neural spellchecker, which improves upon an off-the-shelf spellchecker by incorporating context into spellchecking using a pre-trained neural language model (LM).
\end{itemize}

\section{Background}

\subsection{Transformers}

Transformers \cite{vaswani2017attention} are powerful deep seq2seq architectures that rely heavily on the attention mechanism \cite{bahdanau2015neural, luong2015effective}.
Both the encoder and the decoder of a Transformer are stacks of Transformer blocks, each of which consists of a multi-head self-attention layer followed by a position-wise feed-forward layer, along with residual connection \cite{he2016identity} and layer normalization \cite{ba2016layer}. 
Each decoder block also attends \cite{luong2015effective} to the encoder outputs, in between its self-attention and feed-forward layers.
Each input token embedding in a Transformer is combined with a positional embedding that encodes where the token appeared in the input sequence.

\subsection{Copy-Augmented Transformers}

Copy-augmented Transformers \cite{zhao2019improving} are a class of Transformers that also incorporate an attention-based copying mechanism \cite{gu2016incorporating, see2017get, jia2016data} in the decoder. 
For each output token $y_t$ at output position $t$, the output probability distribution of a copy-augmented Transformer is a mixture of the decoder's generative distribution $p^\mathrm{gen}$ and a copy distribution $p^\mathrm{copy}$, which is defined as an encoder-decoder attention layer that assigns a distribution over tokens appearing in the source sentence.  
By defining a mixture weight parameter $\alpha_t^\mathrm{copy}$ per each decoding step, the output distribution can be compactly represented as follows:
\begin{align}
    p(y_t) &= (1 - \alpha_t^\mathrm{copy}) \cdot p^\mathrm{gen}(y_t) + \alpha_t^\mathrm{copy} \cdot p^\mathrm{copy}(y_t) \label{eqn:copyaugmented}
\end{align}
The mixture weight balances between how likely it is for the model to simply copy a source token, rather than generating a possibly different token.

\subsection{Denoising Autoencoders}

Denoising autoencoders (DAEs) \cite{vincent2008extracting} are a class of neural networks that learns to reconstruct the original input given its noisy version. 
Given an input ${\bf x}$ and a (stochastic) noising function ${\bf x} \mapsto {\tilde{\bf x}}$, the encoder-decoder model of a DAE minimizes the reconstruction loss:
\begin{equation}\label{eqn:dae}
    L({\bf x}, \mathrm{dec}(\mathrm{enc}(\tilde{\bf x})))
\end{equation}
where $L$ is some loss function.

Within the NLP domain, DAEs have been for pre-training in seq2seq tasks that can be cast as a denoising task. 
For example, in GEC, pre-trained DAEs have been used for correcting erroneous sentences \cite{xie2018noising, zhao2019improving}.
Another example is low-resource machine translation (MT) \cite{lample2018phrase}, pre-trained DAEs were used to convert word-by-word translations into natural sentences.

\section{Related Work}

Many recent neural GEC models \cite{junczys2018approaching, lichtarge2018weakly, zhao2019improving} made use of the Transformer \cite{vaswani2017attention} architecture and saw results nearly as good as or better than convolutional \cite{chollampatt2018multilayer, chollampatt2018neural} and recurrent \cite{grundkiewicz2018near, ge2018fluency} architectures.
Recently, \citet{zhao2019improving} further incorporated a copying mechanism \cite{gu2016incorporating,see2017get,jia2016data} to the Transformer, highlighting the fact that most (from 83\% in Lang-8 to 97\% in CoNLL-2013) of the target tokens are exact copies of the corresponding source tokens. 

Several prior results, both early \cite{brockett2006correcting, felice2014generating} and recent \cite{ge2018fluency, xie2018noising, zhao2019improving}, introduced different strategies for generating erroneous text that can in turn be used for model (pre-)training.
One major direction is to introduce an additional ``back-translation'' model \cite{ge2018fluency, xie2018noising}, inspired by its success in NMT \cite{sennrich2016improving}, and let this model learn to generate erroneous sentences from correct ones.
While these back-translation models can learn naturally occurring grammatical errors from the parallel corpora in reverse, they also require relatively large amounts of parallel corpora, which are not readily available in low resource scenarios.
The other direction, which can avoid these issues, is to incorporate a pre-defined noising function, which can generate pre-training data for a denoising task \cite{zhao2019improving}. 
Compared to \cite{zhao2019improving}, our work introduces a noising function that generates more realistic grammatical errors.

\section{Pre-training a Denoising Autoencoder on Realistic Grammatical Errors}\label{sec:pretrain}

Given the relative lack of parallel corpora for the GEC task, it is important to define a realistic pre-training task, from which the learned knowledge can transfer to an improved performance. 

When pre-training a seq2seq model on an auxiliary denoising task, the choice of the noising function is important.
For instance, in low-resource MT, \citet{lample2018unsupervised,lample2018phrase} made use of a noising function that randomly insert/replace/remove tokens or mix up nearby words at uniform probabilities. 
They showed that this approach is effective in translating naive word-by-word translations into correct ones, both because the coverage of word-to-word dictionaries can be limited and because word order is frequently swapped between languages (e.g., going from SVO to SOV). 

In GEC, \citet{zhao2019improving} used a similar noising function to generate a pre-training dataset. 
However, we find that this noising function is less realistic in GEC than in low-resource MT.
For example, randomly mixing up nearby words can be less effective for GEC than for low-resource MT, because word order errors occur less frequently than other major error categories, such as missing punctuations and noun numbers.
Also, replacing a word to any random word in the vocabulary is a less realistic scenario than only replacing it with its associated common error categories, such as prepositions, noun numbers and verb tenses.

To generate realistic pre-training data, we introduce a novel noising function that captures in-domain grammatical errors commonly made by human writers.

\subsection{Constructing Noising Scenarios}\label{sec:noising_scenarios}

We introduce two kinds of noising scenarios, using a token-based approach and a type-based approach.

In the token-based approach, we make use of extracted human edits from annotated GEC corpora, using automated error annotation toolkits such as ERRANT \cite{bryant2017automatic}.
We first take a subset of the training set, preferably one that contains in-domain sentences with high-quality annotations, and using an error annotation toolkit, we collect all edits that occurred in the parallel corpus as well as how often each edit was made. 
We then take edits that occur in for at least $k$ times, where $k$ is a pre-defined threshold (we fix $k=4$ in our experiments), in order to prevent overfitting to this (possibly small) subset.
These extracted edits include errors commonly made by human writers, including missing punctuations (e.g., adding a comma), preposition errors (e.g., \textit{of} $\rightarrow$ \textit{at}), and verb tenses (e.g., \textit{has} $\rightarrow$ \textit{have}). 
As a result, we obtain an automatically constructed dictionary of common edits made by human annotators on the in-domain training set.
Then, we can define a realistic noising scenario by randomly applying these human edits, in reverse, to a grammatically correct sentence.

In the type-based approach, we also make use of \textit{a priori} knowledge and construct a noising scenario based on token types, including prepositions, nouns, and verbs. 
For each token type, we define a noising scenario based on commonly made errors associated with that token type, but without changing the type of the original token.
In particular, we replace prepositions with other prepositions, nouns with their singular/plural version, and verbs with one of their inflected versions.
This introduces another set of realistic noising scenarios, thereby increasing the coverage of the resulting noising function.

\subsection{Generating Pre-training Data}\label{sec:generate_data}

Our goal is to come up with an error function that introduces grammatical errors that are commonly made by human writers in a specific setting (in this case, personal essays written by English students). 
Given sets of realistic noising scenarios, we can generate large amounts of erroneous sentences from high-quality English corpora, such as the Project Gutenberg corpus \cite{lahiri2014complexity} and Wikipedia \cite{merity2016pointer}.

We first check if a token exists in the dictionary of token edits. If it does, a token-based error is generated with the probability of 0.9. Specifically, the token is replaced by one of the associated edits with the probabilities proportional to the frequency of each edit.
For example, the token \textit{for} may be replaced with \textit{during}, \textit{in}, \textit{four}, and also \textit{for} (coming from a \textit{noop} edit). 

If a token is not processed through the token-based scenario, we then examine if it belongs to one of the pre-defined token types: in our case, we use prepositions, nouns, and verbs. 
If the token belongs to one such type, we then apply the corresponding noising scenario.

\section{Sequential Transfer Learning}\label{sec:transfer}

\subsection{Transferring Pre-trained DAE Weights}\label{sec:train}

As discussed in \cite{zhao2019improving}, an important benefit of pre-training a DAE is that it provides good initial values for both the encoder and the decoder weights in the seq2seq model. 
Given a pre-trained DAE, we initialize our seq2seq GEC model using the learned weights of the DAE and train on all available parallel training corpora with smaller learning rates.
This model transfer approach \cite{wang2015transfer} can be viewed as a (relatively simple) version of sequential transfer learning \cite{ruder2019neural}.

\subsection{Adaptation by Fine-tuning}\label{sec:finetune}

As noted in \cite{junczys2018approaching}, the distribution of grammatical errors occurring in text can differ across the domain and content of text. 
For example, a Wikipedia article introducing a historical event may involve more rare words than a personal essay would. 
The distribution can also be affected significantly by the writer's style and proficiency, as well as the annotator's preferred style of writing (e.g., British vs. American styles, synonymous word choices, and Oxford commas). 

In this work, given that the primary source of evaluation are personal essays at various levels of English proficiency -- in particular the W\&I+LOCNESS dataset \cite{yannakoudakis2018developing} -- we adapt our trained models to such characteristics of the test set by fine-tuning the model only on the training portion of W\&I, which largely matches the domain of the development and test sets.\footnote{This is analogous to the NUCLE dataset matching ``perfectly'' with the CoNLL dataset, as noted in \cite{junczys2018approaching}.}
Similar to our training step in \S\ref{sec:train}, we use (even) smaller learning rates.
Overall, this sequential transfer learning framework can also be viewed as an alternative to oversampling in-domain data sources, as proposed in \cite{junczys2018approaching}.

\section{A Context-Aware Neural Spellchecker}\label{sec:spellchecker}

Many recent GEC systems include an off-the-shelf spellchecker, such as the open-source package \texttt{enchant} \cite{sakaguchi2017grammatical, junczys2018approaching} and Microsoft's Bing spellchecker \cite{ge2018fluency, ge2018reaching}. 
While the idea of incorporating context into spellchecking has been repeatedly discussed in the literature \cite{flor2012using, chollampatt2017connecting}, popular open-sourced spellcheckers such as \texttt{hunspell} primarily operate at the word level. 
This fundamentally limits their capacity, because it is often difficult to find which word is intended for without context.
For example, given the input sentence \textit{This is an esay about my favorite sport.}, \texttt{hunspell} invariably suggests \textit{easy} as its top candidate for \textit{esay}, which should actually be corrected as \textit{essay}.

Our spellchecker incorporates context to \texttt{hunspell} using a pre-trained neural language model (LM). 
Specifically, we re-rank the top candidates suggested by \texttt{hunspell} through feeding each, along with the context, to the neural LM and scoring them.

\begin{table}[t]
    \centering
    \begin{tabular}{cccc}
        \Xhline{1.1pt}
        \textbf{Source} & \textbf{Public?} & \textbf{\# Sent.} & \textbf{\# Annot.} \\ \hline
        Gutenberg & Yes & 11.6M & n/a \\ 
        Tatoeba & Yes & 1.17M & n/a \\ 
        WikiText-103 & Yes & 3.93M & n/a \\ \hline 
        FCE & Yes & 33.2K & 1 \\ 
        Lang-8 & Yes & 1.04M & 1-8 \\
        NUCLE & Yes & 57.2K & 1 \\ 
        W\&I-Train & Yes & 34.3K & 1 \\ \hline 
        W\&I+L-Dev & Yes & 4.38K & 1 \\ 
        \hline
        W\&I+L-Test & Yes & 4.48K & 5 \\ 
        \Xhline{1.1pt}
    \end{tabular}
    \caption{Summary of datasets. 
    The first three datasets are unannotated English corpora, from which we generate parallel data for pre-training using a pre-defined noising function.
    }
    \label{tbl:datasets}
\end{table}

\begin{table*}[t]
    \centering
    \begin{tabular}{cccc}
        \Xhline{1.1pt}
         & {\bf Restricted (\S\ref{sec:track1})} & {\bf Low Resource (\S\ref{sec:track3})} & {\bf CoNLL-2014 (\S\ref{sec:track0})} \\ \hline
        {\bf Error Extraction} & W\&I Train & W\&I+L Dev-3K & NUCLE \\ \hline
        {\bf Pre-training} & \multicolumn{3}{c}{Gutenberg, Tatoeba, WikiText-103} \\ \hline
        \multirow{2}{*}{\bf Training} & FCE, Lang-8, NUCLE, & \multirow{2}{*}{W\&I+L Dev-3K} & \multirow{2}{*}{FCE, Lang-8, NUCLE} \\
         & W\&I Train & & \\
        \hline
        {\bf Fine-tuning} & W\&I Train & n/a & NUCLE \\ \hline
        {\bf Validation} & W\&I+L Dev & W\&I+L Dev-1K & CoNLL-2013 \\ \hline
        {\bf Test} & W\&I+L Test & W\&I+L Test & CoNLL-2014 \\
        \Xhline{1.1pt}
        \end{tabular}
    \caption{Datasets used for each set of results. 
    For the W\&I+L development set, Dev-3K and Dev-1K respectively indicate a 3:1 train-test random split of the development set, such that the original proportions of English proficiency (A, B, C, N) are kept the same in each split.
    See Table \ref{tbl:datasets} for more information about each dataset.
    }
    \label{tbl:data_usage}
\end{table*}

\begin{figure*}[t]
    \centering
    \includegraphics[width=16cm]{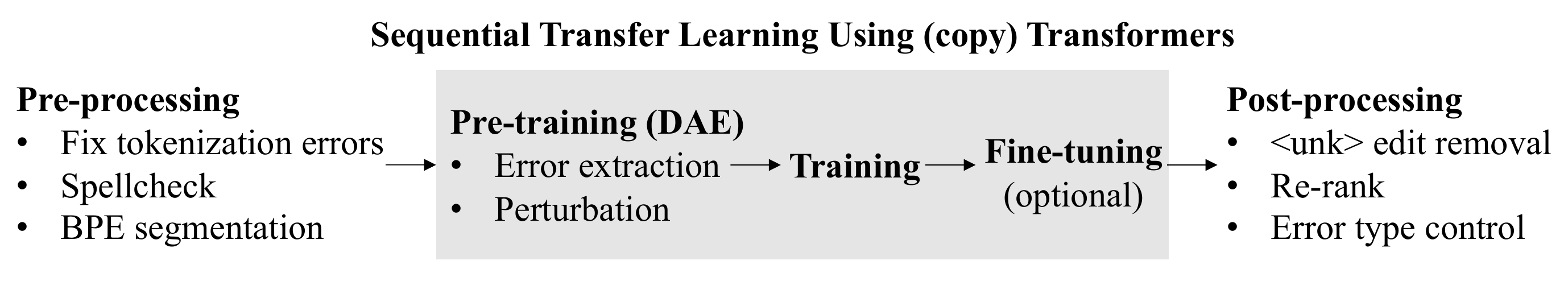}
    \caption{Overall pipeline for our approach. 
    Gray shaded box includes the training steps for a seq2seq model. 
}
    \label{fig:pipeline}
\end{figure*}

\section{Experiments}

Throughout our experiments, we use \texttt{fairseq}\footnote{\url{https://github.com/pytorch/fairseq}} \cite{ott2019fairseq}, a publicly available sequence-to-sequence modeling toolkit based on PyTorch \cite{paszke2017automatic}. 
Specifically, we take \texttt{fairseq-0.6.1} and add our own implementations of a copy-augmented transformer model as well as several GEC-specific auxiliary losses.

\subsection{Datasets \& Setups}\label{sec:datasets}

In Table \ref{tbl:datasets}, we summarize all relevant data sources, their sizes, whether they are public, and the number of annotators.

For pre-training, we use the Gutenberg dataset \cite{lahiri2014complexity}, the Tatoeba\footnote{\url{https://tatoeba.org/eng/downloads}} dataset, and the WikiText-103 dataset \cite{merity2016pointer}. 
We learned through initial experiments that the quality of pre-training data is crucial to the final model's performance, because our DAE model assumes \S\ref{sec:pretrain} that these unannotated corpora contain little grammatical errors. 
Our choice of corpora is based on both the quality and diversity of text: Gutenberg contains clean novel writings with minimal grammatical errors, Tatoeba contains colloquial sentences used as sample sentences in dictionaries, and WikiText-103 contains ``Good'' and ``Featured'' articles from Wikipedia.
Our final pre-training data is a collection of 45M (perturbed, correct) sentence pairs based on these datasets, with our noising approach (\S\ref{sec:pretrain}) applied multiple times to each dataset to approximately balance data from each source (1x Gutenberg, 12x Tatoeba, and 5x WikiText-103). 

Our default setup is the ``Restricted Track'' scenario (\S\ref{sec:track1}) for the BEA 2019 Shared Task, where we use four data sources: the FCE dataset \cite{bea2019}, the Lang-8 dataset\footnote{As in previous results, we remove all duplicates but take multiple annotations (if available) the Lang-8 dataset, leaving only 575K parallel examples.} \cite{mizumoto2011mining,tajiri2012tense}, the NUCLE (v3.3) dataset \cite{dahlmeier2013building}, and the newly released Write \& Improve and LOCNESS (W\&I+L) datasets \cite{yannakoudakis2018developing}.\footnote{See Appendix \ref{app:exploratory} for an exploratory data analysis.}
For the ``Low Resource Track'' (\S\ref{sec:track3}), we use a 3:1 train-test random split of the W\&I+L development set, keeping the proportions of proficiency levels the same. 
In both tracks, we report our final results on the W\&I+L test set, which contains 5 annotations.
Further, because the W\&I+L dataset is relatively new, we also include results on the CoNLL-2014 \cite{ng2014conll} dataset, with and without using the W\&I+L dataset during training  (\S\ref{sec:track0}).
In Table \ref{tbl:data_usage}, we summarize which datasets were used in each setup.

\subsection{Pre-processing}

As part of pre-processing, we first fix minor tokenization issues in the dataset using regular expressions.
We use spaCy v1.9 \cite{honnibal2017spacy} to make tokenization consistent with the final evaluation module (ERRANT).

This tokenized input is then fed to our context-aware neural spellchecker (\S\ref{sec:spellchecker}).
For the neural LM, we use a gated convolutional neural network language model \cite{dauphin2017language} pre-trained on WikiText-103 \cite{merity2016pointer}. 

During spellchecking, we also found it beneficial to fix casing errors within our context-aware spellchecking process. 
To fix case errors, we extract a list of words used in the capital form much more than their lower-case version (more than 99 times) in WikiText-103 \cite{merity2016pointer}. 
We then include a capitalized version of the word as a candidate in the LM re-scoring process if it appears in its capitalized form is in the extracted list of common capital words. 

Before feeding spellchecked text into our seq2seq model, we apply byte-pair encoding (BPE) \cite{sennrich2016neural} using SentencePiece \cite{kudo2018sentencepiece}. 
We first train a SentencePiece model with 32K vocabulary size on the original Gutenberg corpus, and apply this model to all input text to the model. 
This allows us to avoid \texttt{<unk>} tokens in most training and validation sets, including the W\&I+L development set.

\subsection{Model \& Training Details}

Throughout our experiments, we use two variants of the Transformer model: the ``vanilla'' Transformer \cite{vaswani2017attention} and the copy-augmented Transformer \cite{zhao2019improving}. 
We use two configurations for the vanilla Transformer: a \textit{base} model with 6 blocks of 512-2048 units with 8 attention heads, and a \textit{large} model with 6 blocks of 1024-4096 units with 16 attention heads and pre-attention layer normalization.
We only use the large model for Restricted Track (\S\ref{sec:track1}) and for the CoNLL-2014 comparison (\S\ref{sec:track0}). 
For the \textit{copy}-augmented Transformer, we follow the default configuration from \cite{zhao2019improving}: 6 blocks of 512-4096 units with 8 attention heads, along with an 8-head copy attention layer. 
For each model configuration, we train two independent models using different seeds.

Our model training is a three-stage process, as illustrated in Figure \ref{fig:pipeline}: DAE pre-training, training, and fine-tuning, except in Low Resource Track where there is no fine-tuning data (see Table \ref{tbl:data_usage}). 
At each step, we train a model until its ERRANT score on the development set reaches convergence, and use the learned weights as initial values for the next step. 
In all training steps, we used the Adam \cite{kingma2014adam} optimizer.

Our final model is an ensemble among the different model configurations and seeds. 
Among the six (four for Low Resource Track) best models, we greedily search for the best combination, starting with the best-performing single model. 

\subsection{Post-processing}

Our post-processing phase involves three steps.
First, we find any \texttt{<unk>} tokens found in the original input text, and using ERRANT, we remove any edits associated with the token.
Next, since many of the model's corrections can still be unnatural, if not incorrect, we re-rank candidate corrections within each sentence using a pre-trained neural LM \cite{dauphin2017language}. 
Specifically, we remove any combination of up to 7 edits per sentence, and choose the combination that yields the highest LM score. 
Finally, we noticed that, as in many previous results, our neural system performs well on some error categories (e.g., M:PUNCT) but poorly on others (e.g., R:OTHER). 
Because ERRANT provides a fine-grained analysis of model performance based on error types, we found it beneficial to remove edits belonging to certain categories in which the model performs too poorly. 
Given our final model, we randomly remove all edits from a subset of (at most $N$) categories for repeated steps, and choose to remove the subset of error categories that gave the highest score on the development set.

\begin{table*}[t]
    \centering
    \begin{tabular}{l|cccc|cccc}
        \Xhline{1.1pt}
        \multirowcell{2}{\textbf{Steps}} & \multicolumn{4}{c|}{\textbf{W\&I+L Dev}} & \multicolumn{4}{c}{\textbf{W\&I+L Test}} \\
         & \textbf{P} & \textbf{R} & \textbf{F0.5} & \textbf{$\Delta$} & \textbf{P} & \textbf{R} & \textbf{F0.5} & \textbf{$\Delta$} \\ \hline
        Spellcheck              & 59.28 & \;\;5.27 & {\bf 19.43} &  n/a   & 68.77 & 10.55 & {\bf 32.69 }&   n/a  \\ \hline
        + DAE Pre-train   & 48.58 & 24.92 & {\bf 40.82} & +21.39 & 58.33 & 44.20 & {\bf 54.82} & +22.13 \\ 
        + Train          & 54.30 & 28.67 & {\bf 46.07} &  +\;\;5.25 & 66.05 & 50.72 & {\bf 62.28} &  +\;\;7.46 \\ 
        + Fine-tune    & 54.34 & 32.15 & {\bf 47.75} &  +\;\;1.68 & 66.02 & 53.41 & {\bf 63.05} &  +\;\;0.77 \\ \hline
        + Ensemble  (5)  & 63.54 & 31.48 & {\bf 52.79} & +\;\;5.04 & 76.19 & 50.25 & {\bf 69.06} & +\;\;6.01 \\ 
        \Xhline{1.1pt}
    \end{tabular} 
    \caption{
    ACL 2019 BEA Workshop \textbf{Restricted Track} results. 
    For each training step, we only list results from the model configuration that achieved the best $F_{0.5}$ test set score.
    All evaluation is done using ERRANT's span-based correction scorer.
    Pre-processing and post-processing are included in the first step and last steps, respectively.
    }
    \label{tbl:track1}
\end{table*}

\begin{table*}[t]
    \centering
    \begin{tabular}{l|cccc|cccc}
        \Xhline{1.1pt}
        \multirowcell{2}{\textbf{Steps}} & \multicolumn{4}{c|}{\textbf{W\&I+L Dev-1K}} & \multicolumn{4}{c}{\textbf{W\&I+L Test}} \\
         & \textbf{P} & \textbf{R} & \textbf{F0.5} & \textbf{$\Delta$} & \textbf{P} & \textbf{R} & \textbf{F0.5} & \textbf{$\Delta$} \\ \hline
        Spellcheck             & 61.88 & \;\;5.29 & {\bf 19.72} &  n/a   & 68.77 & 10.55 & {\bf 32.69} &   n/a  \\ \hline 
        + DAE Pre-train  & 46.26 & 19.84 & {\bf 36.53} & +16.81 & 57.14 & 37.46 & {\bf 51.71} & +19.02 \\ 
        + Train     & 47.97 & 30.91 & {\bf 43.20} & +\;\;6.67 & 58.60 & 47.47 & {\bf 55.98} & +\;\;4.27 \\  \hline
        + Ensemble (4) & 58.89 & 26.68 & {\bf 47.02} & +\;\;5.75 & 69.69 & 41.76 & {\bf 61.47} & +\;\;5.49 \\ 
        \Xhline{1.1pt}
    \end{tabular} 
    \caption{
    ACL 2019 BEA Workshop \textbf{Low Resource Track} results. 
    For each training step, we only list results from the model configuration that achieved the best $F_{0.5}$ test set score.
    All evaluation is done using ERRANT's span-based correction scorer. Note that 3K examples from the W\&I+Locness development set (``W\&I+L Dev-3K'') were used for the training step and is excluded during evaluation. 
    Pre-processing and post-processing are included in the first step and last steps, respectively.
    }
    \label{tbl:track3}
\end{table*}

\subsection{Restricted Track Results}\label{sec:track1}

In Table \ref{tbl:track1}, we summarize our results on Restricted Track. 
The results illustrate that each step in our approach substantially improves upon the previous model, both on the W\&I+L development and test sets. 
We highlight that our pre-training step with realistic human errors already gets us at a 54.82 $F_{0.5}$ score on span-based correction in ERRANT for the test set, even though we only indirectly used the W\&I training set for error extraction and no other parallel corpora. 
This suggests that pre-training on a denoising task with realistic and common errors can already lead to a decent GEC system.

Our final ensemble model is a combination of five independent models -- one base model, two large models, and two copy-augmented models -- achieving 69.06 $F_{0.5}$ score on the test set.

\subsection{Low Resource Track Results}\label{sec:track3}

In Table \ref{tbl:track3}, we summarize our results on Low Resource Track. 
Similar to Restricted Track, each step in our approach improves upon the previous model significantly, and despite the lack of parallel data (3K for training, 1K for validation), our pre-training step already gets us at 51.71 $F_{0.5}$ score on the test set. 
Compared to Restricted Track, the only difference in pre-training is that the reverse dictionary for the noising function was constructed using much fewer parallel data (3K), but we see that this amount of parallel data is already enough to get within 3 points of our pre-trained model in Restricted Track.

Our final model is an ensemble of two independent models -- one base model and one copy model -- achieving 61.47 $F_{0.5}$ score on the test set.

\subsection{CoNLL-2014 Results}\label{sec:track0}

\begin{table*}[t]
    \centering
    \begin{tabular}{cccccc}
        \Xhline{1.1pt}
        \multirowcell{2}{\textbf{Models}} & \multirowcell{2}{\textbf{Pre-training}} & \multirowcell{2}{\textbf{W\&I+L}} & \multicolumn{3}{c}{\textbf{CoNLL-2014}} \\
         & & & \textbf{P} & \textbf{R} & \textbf{F0.5} \\ \hline
        \multicolumn{6}{c}{\textbf{Our Models}} \\ \hline
        \makecell[c]{Transformers \\ (Vanilla + Copy-Aug.)} & DAE with Realistic Errors & \makecell{No \\ Yes} & \makecell{71.11 \\ 74.76} & \makecell{32.56 \\ 34.05} & \makecell{\textbf{57.50} \\ \textbf{60.33}} \\ \hline
        \multicolumn{6}{c}{\textbf{Previous Results}} \\ \hline
        Copy-Aug. Transformers  & DAE with Random Errors               & No  & 71.57 & 38.65 & \textbf{61.15} \\ 
        Transformers                 & Revisions + Round-Trip Translations & No  & 66.70 & 43.90 & 60.40 \\
        ConvS2S + QE                 & LM (Decoder Pre-training)            & No  & n/a   & n/a   & 56.52 \\ 
        SMT + BiGRU                  & LM (Ensemble Decoding)               & No  & 66.77 & 34.49 & 56.25 \\ 
        \Xhline{1.1pt}
    \end{tabular} 
    \caption{Results on CoNLL-2014 as point of comparison. ``W\&I+L'' indicates whether the approach made use of the (newly released) W\&I+L dataset. Evaluation is done using the MaxMatch ($M^2$) scorer, rather than ERRANT. Pre-processing \& post-processing are included before the first step and after the last step, respectively. See \S\ref{sec:track0} for details and references.}
    \label{tbl:track0}
\end{table*}

In Table \ref{tbl:track0}, we show the performance of our approach on the CoNLL-2014 \cite{ng2014conll} dataset, with and without the newly released W\&I+L dataset.\footnote{See Appendix \ref{app:training_detail} for a step-by-step training progress.} 
We also list some of the state-of-the-art\footnote{\url{http://nlpprogress.com/english/grammatical_error_correction.html}.} results prior to the shared task: copy-augmented Transformers pre-trained on random error denoising \cite{zhao2019improving}, Transformers pre-trained on Wikipedia revisions and round-trip translations \cite{lichtarge2019corpora}, hybrid statistical and neural machine translation systems \cite{junczys2018approaching}, and convolutional seq2seq models with quality estimation \cite{chollampatt2018neural}.

The results show that our approach is competitive with some of the recent state-of-the-art results that achieve around 56 MaxMatch ($M^2$) scores and further achieves 60+ $M^2$ score when the W\&I+L dataset is used. 
This illustrates that our approach can also achieve a ``near human-level performance'' \cite{grundkiewicz2018near}.
We also note that the 60.33 $M^2$ score was obtained by the final ensemble model from \S\ref{sec:track1}, which includes a fine-tuning step to the W\&I model. 
This suggests that ``overfitting'' to the W\&I dataset does not necessarily imply a reduced performance on an external dataset such as CoNLL-2014.

\section{Analysis \& Discussion} 

\subsection{Error Analysis}

Here, we give an analysis of our model's performance on some of the major ERRANT error categories on the W\&I test set. Detailed information is available in Tabel \ref{tbl:error_categories}.
We observe that our model performs well on syntax relevant error types, i.e., subject-verb agreement (VERB:SVA) (84.09 $F_{0.5}$), noun numbers (NOUN:NUM) (72.19), and prepositions (PREP) (64.27), all of which are included as part of our type-based error generation in the pre-training data (\S\ref{sec:generate_data}). 
Our model also achieves 77.26 on spelling errors (SPELL) and 75.83 on orthographic errors (ORTH), both of which are improvements made mostly by our context-aware neural spellchecker. 
Our model also achieves 77.86 on punctuation errors (PUNCT), which happen to be the most common error category in the W\&I+L dataset.
This may be due to both our use of extracted errors from the W\&I dataset during pre-training and our fine-tuning step. 
Finally, we find it challenging to match human annotators' ``naturalness'' edits, such as VERB (26.76), NOUN (41.67), and OTHER (36.53). 
This is possibly due to the variability in annotation styles and a lack of large training data with multiple human annotations.

\subsection{Effect of Realistic Error Generation}\label{sec:realistic}

\begin{table}
\centering
\begin{tabular}{l>{\bfseries}ccc}
\Xhline{1.1pt}
\textbf{Step} & \textbf{Ours} & \textbf{Random} & \textbf{$\Delta$} \\
\hline
DAE     & 54.82 & 32.25 & +22.57 \\
+ Train & 62.28 & 57.00 & +\; 5.28 \\
+ Fine-tune & 63.05 & 60.22 & +\; 2.83 \\
\Xhline{1.1pt}
\end{tabular}

\caption{Comparison of realistic and random error generation on Restricted Track. $\Delta$ means the difference between \textbf{Ours} and \textbf{Random}.}
\label{tbl:random-track1}
\end{table}

\begin{table}
\centering

\begin{tabular}{l>{\bfseries}ccc}
\Xhline{1.1pt}
\textbf{Step} & \textbf{Ours} & \textbf{Random} & $\Delta$ \\
\hline
DAE & 51.71 & 32.01 & +19.70 \\
+ Train & 55.98 & 35.44 & +20.54 \\
\Xhline{1.1pt}
\end{tabular}

\caption{Comparison of realistic and random error generation on Low Resource Track. $\Delta$ means the difference between \textbf{Ours} and \textbf{Random}.}
\label{tbl:random-track3}
\end{table}

To see how effective our realistic error based pre-training is, we compare it with \cite{zhao2019improving}'s method.
According to them, random insertion, deletion, and substitution occur with the probability of 0.1 at every word, and words are re-ordered with a certain probability.
As seen in Table \ref{tbl:random-track1} and \ref{tbl:random-track3}, our pre-training method outperforms the random based one in both Restricted and Low Resource Tracks by 22.57 and 19.70, respectively.
And it remains true for each step of the following transfer learning.
The performance gap, however, decreases to 5.3 after training and to 3.2 after fine-tuning in Restricted Track.
On the other hand, the gap in Low Resource Track slightly increases to 20.54 after training.
This leads to the conclusion that our pre-training functions as proxy for training, for our generated errors resemble the human errors in the training data more than the random errors do.

\subsection{Effect of Context-Aware Spellchecking}

We further investigate the effects of incorporating context and fixing casing errors to the off-the-shelf \texttt{hunspell}, which we consider as a baseline.
We test three spellchecker variants: \texttt{hunspell}, \texttt{hunspell} using a neural LM, and our final spellchecker model.

On the original W\&I+L test set, our LM-based approach improves upon the ERRANT F0.5 score by 5.07 points, and fixing casing issues further improves this score by 4.02 points. 
As a result, we obtain 32.69 F0.5 score just by applying our context-aware spellchecker model.

\begin{table}[t]
    \centering
    \begin{tabular}{ccc>{\bfseries}cc}
        \Xhline{1.1pt}
        \multirowcell{2}{\textbf{Spellchecker}} & \multicolumn{4}{c}{\textbf{W\&I+L Test}} \\
         & \textbf{P} & \textbf{R} & \textbf{F0.5} & \textbf{$\Delta$} \\ \hline
        Hunspell      & 53.59 & 7.29  & 23.60 & n/a    \\ 
        Hunspell + LM     & 65.14 & 8.85 & 28.67 & +5.07    \\
        Ours          & 68.77 & 10.55 & 32.69 & +4.02    \\ 
        \Xhline{1.1pt}
    \end{tabular} 
    \caption{Effect of incorporating context into a standard spellchecker.} 
    \label{tbl:spellcheck}
\end{table}

\section{Conclusion \& Future Work}

We introduced a neural GEC system that leverages pre-training using realistic errors, sequential transfer learning, and context-aware spellchecking with a neural LM. 
Our system achieved competitive results on the newly released W\&I+L dataset in both standard and low-resource settings.

There are several interesting future directions following our work. 
One is to extend sentence-level GEC systems to multi-sentence contexts, for example by including the previous sentence, to better cope with complex semantic errors such as collocation.
Because the W\&I+L dataset is also a collection of (multi-)paragraph essays, adding multi-sentence contexts can improve these GEC systems.
Also, to better understand the role of several components existing in modern GEC systems, it is important to examine which components are more necessary than others.

\bibliography{acl2019}

\begin{thebibliography}{49}
\expandafter\ifx\csname natexlab\endcsname\relax\def\natexlab#1{#1}\fi

\bibitem[{Ba et~al.(2016)Ba, Kiros, and Hinton}]{ba2016layer}
Jimmy~Lei Ba, Jamie~Ryan Kiros, and Geoffrey~E Hinton. 2016.
\newblock Layer normalization.
\newblock \emph{arXiv preprint arXiv:1607.06450}.

\bibitem[{Bahdanau et~al.(2015)Bahdanau, Cho, and Bengio}]{bahdanau2015neural}
Dzmitry Bahdanau, Kyunghyun Cho, and Yoshua Bengio. 2015.
\newblock Neural machine translation by jointly learning to align and
  translate.
\newblock In \emph{ICLR}.

\bibitem[{Brockett et~al.(2006)Brockett, Dolan, and
  Gamon}]{brockett2006correcting}
Chris Brockett, William~B Dolan, and Michael Gamon. 2006.
\newblock Correcting esl errors using phrasal smt techniques.
\newblock In \emph{Proceedings of the 21st International Conference on
  Computational Linguistics and the 44th annual meeting of the Association for
  Computational Linguistics}, pages 249--256. Association for Computational
  Linguistics.

\bibitem[{Bryant et~al.(2019)Bryant, Felice, Andersen, and Briscoe}]{bea2019}
Christopher Bryant, Mariano Felice, {\O}istein~E. Andersen, and Ted Briscoe.
  2019.
\newblock {The BEA-2019 Shared Task on Grammatical Error Correction}.
\newblock In \emph{Proceedings of the 14th Workshop on Innovative Use of NLP
  for Building Educational Applications}. Association for Computational
  Linguistics.

\bibitem[{Bryant et~al.(2017)Bryant, Felice, and Briscoe}]{bryant2017automatic}
Christopher Bryant, Mariano Felice, and Edward~John Briscoe. 2017.
\newblock Automatic annotation and evaluation of error types for grammatical
  error correction.
\newblock In \emph{ACL}.

\bibitem[{Cho et~al.(2014)Cho, van Merrienboer, Gulcehre, Bahdanau, Bougares,
  Schwenk, and Bengio}]{cho2014learning}
Kyunghyun Cho, Bart van Merrienboer, Caglar Gulcehre, Dzmitry Bahdanau, Fethi
  Bougares, Holger Schwenk, and Yoshua Bengio. 2014.
\newblock Learning phrase representations using rnn encoder--decoder for
  statistical machine translation.
\newblock In \emph{EMNLP}.

\bibitem[{Chollampatt and Ng(2017)}]{chollampatt2017connecting}
Shamil Chollampatt and Hwee~Tou Ng. 2017.
\newblock Connecting the dots: Towards human-level grammatical error
  correction.
\newblock In \emph{Proceedings of the 12th Workshop on Innovative Use of NLP
  for Building Educational Applications}, pages 327--333.

\bibitem[{Chollampatt and Ng(2018{\natexlab{a}})}]{chollampatt2018multilayer}
Shamil Chollampatt and Hwee~Tou Ng. 2018{\natexlab{a}}.
\newblock \href
  {https://www.aaai.org/ocs/index.php/AAAI/AAAI18/paper/view/17308} {A
  multilayer convolutional encoder-decoder neural network for grammatical error
  correction}.
\newblock In \emph{Proceedings of the Thirty-Second {AAAI} Conference on
  Artificial Intelligence, (AAAI-18), the 30th innovative Applications of
  Artificial Intelligence (IAAI-18), and the 8th {AAAI} Symposium on
  Educational Advances in Artificial Intelligence (EAAI-18), New Orleans,
  Louisiana, USA, February 2-7, 2018}, pages 5755--5762.

\bibitem[{Chollampatt and Ng(2018{\natexlab{b}})}]{chollampatt2018neural}
Shamil Chollampatt and Hwee~Tou Ng. 2018{\natexlab{b}}.
\newblock Neural quality estimation of grammatical error correction.
\newblock In \emph{Proceedings of the 2018 Conference on Empirical Methods in
  Natural Language Processing}, pages 2528--2539.

\bibitem[{Dahlmeier et~al.(2013)Dahlmeier, Ng, and Wu}]{dahlmeier2013building}
Daniel Dahlmeier, Hwee~Tou Ng, and Siew~Mei Wu. 2013.
\newblock Building a large annotated corpus of learner english: The nus corpus
  of learner english.
\newblock In \emph{Proceedings of the eighth workshop on innovative use of NLP
  for building educational applications}, pages 22--31.

\bibitem[{Dauphin et~al.(2017)Dauphin, Fan, Auli, and
  Grangier}]{dauphin2017language}
Yann~N. Dauphin, Angela Fan, Michael Auli, and David Grangier. 2017.
\newblock \href {http://proceedings.mlr.press/v70/dauphin17a.html} {Language
  modeling with gated convolutional networks}.
\newblock In \emph{Proceedings of the 34th International Conference on Machine
  Learning}, volume~70 of \emph{Proceedings of Machine Learning Research},
  pages 933--941, International Convention Centre, Sydney, Australia. PMLR.

\bibitem[{Felice and Yuan(2014)}]{felice2014generating}
Mariano Felice and Zheng Yuan. 2014.
\newblock \href {https://doi.org/10.3115/v1/E14-3013} {Generating artificial
  errors for grammatical error correction}.
\newblock In \emph{Proceedings of the Student Research Workshop at the 14th
  Conference of the European Chapter of the Association for Computational
  Linguistics}, pages 116--126, Gothenburg, Sweden. Association for
  Computational Linguistics.

\bibitem[{Flor and Futagi(2012)}]{flor2012using}
Michael Flor and Yoko Futagi. 2012.
\newblock On using context for automatic correction of non-word misspellings in
  student essays.
\newblock In \emph{Proceedings of the seventh workshop on building educational
  applications Using NLP}, pages 105--115. Association for Computational
  Linguistics.

\bibitem[{Ge et~al.(2018{\natexlab{a}})Ge, Wei, and Zhou}]{ge2018fluency}
Tao Ge, Furu Wei, and Ming Zhou. 2018{\natexlab{a}}.
\newblock Fluency boost learning and inference for neural grammatical error
  correction.
\newblock In \emph{Proceedings of the 56th Annual Meeting of the Association
  for Computational Linguistics (Volume 1: Long Papers)}, volume~1, pages
  1055--1065.

\bibitem[{Ge et~al.(2018{\natexlab{b}})Ge, Wei, and Zhou}]{ge2018reaching}
Tao Ge, Furu Wei, and Ming Zhou. 2018{\natexlab{b}}.
\newblock Reaching human-level performance in automatic grammatical error
  correction: An empirical study.
\newblock \emph{arXiv preprint arXiv:1807.01270}.

\bibitem[{Gehring et~al.(2017)Gehring, Auli, Grangier, Yarats, and
  Dauphin}]{gehring2017convolutional}
Jonas Gehring, Michael Auli, David Grangier, Denis Yarats, and Yann~N Dauphin.
  2017.
\newblock Convolutional sequence to sequence learning.
\newblock In \emph{Proceedings of the 34th International Conference on Machine
  Learning-Volume 70}, pages 1243--1252. JMLR. org.

\bibitem[{Grundkiewicz and Junczys-Dowmunt(2018)}]{grundkiewicz2018near}
Roman Grundkiewicz and Marcin Junczys-Dowmunt. 2018.
\newblock Near human-level performance in grammatical error correction with
  hybrid machine translation.
\newblock In \emph{Proceedings of the 2018 Conference of the North American
  Chapter of the Association for Computational Linguistics: Human Language
  Technologies, Volume 2 (Short Papers)}, volume~2, pages 284--290.

\bibitem[{Gu et~al.(2016)Gu, Lu, Li, and Li}]{gu2016incorporating}
Jiatao Gu, Zhengdong Lu, Hang Li, and Victor~O.K. Li. 2016.
\newblock \href {https://doi.org/10.18653/v1/P16-1154} {Incorporating copying
  mechanism in sequence-to-sequence learning}.
\newblock In \emph{Proceedings of the 54th Annual Meeting of the Association
  for Computational Linguistics (Volume 1: Long Papers)}, pages 1631--1640,
  Berlin, Germany. Association for Computational Linguistics.

\bibitem[{He et~al.(2016)He, Zhang, Ren, and Sun}]{he2016identity}
Kaiming He, Xiangyu Zhang, Shaoqing Ren, and Jian Sun. 2016.
\newblock Identity mappings in deep residual networks.
\newblock In \emph{European conference on computer vision}, pages 630--645.
  Springer.

\bibitem[{Honnibal and Montani(2017)}]{honnibal2017spacy}
Matthew Honnibal and Ines Montani. 2017.
\newblock spacy 2: Natural language understanding with bloom embeddings,
  convolutional neural networks and incremental parsing.
\newblock \emph{To appear}.

\bibitem[{Jia and Liang(2016)}]{jia2016data}
Robin Jia and Percy Liang. 2016.
\newblock \href {https://doi.org/10.18653/v1/P16-1002} {Data recombination for
  neural semantic parsing}.
\newblock In \emph{Proceedings of the 54th Annual Meeting of the Association
  for Computational Linguistics (Volume 1: Long Papers)}, pages 12--22, Berlin,
  Germany. Association for Computational Linguistics.

\bibitem[{Junczys-Dowmunt and Grundkiewicz(2016)}]{junczys2016phrase}
Marcin Junczys-Dowmunt and Roman Grundkiewicz. 2016.
\newblock \href {https://doi.org/10.18653/v1/D16-1161} {Phrase-based machine
  translation is state-of-the-art for automatic grammatical error correction}.
\newblock In \emph{Proceedings of the 2016 Conference on Empirical Methods in
  Natural Language Processing}, pages 1546--1556, Austin, Texas. Association
  for Computational Linguistics.

\bibitem[{Junczys-Dowmunt et~al.(2018)Junczys-Dowmunt, Grundkiewicz, Guha, and
  Heafield}]{junczys2018approaching}
Marcin Junczys-Dowmunt, Roman Grundkiewicz, Shubha Guha, and Kenneth Heafield.
  2018.
\newblock Approaching neural grammatical error correction as a low-resource
  machine translation task.
\newblock In \emph{Proceedings of the 2018 Conference of the North American
  Chapter of the Association for Computational Linguistics: Human Language
  Technologies, Volume 1 (Long Papers)}, volume~1, pages 595--606.

\bibitem[{Kingma and Ba(2015)}]{kingma2014adam}
Diederik~P Kingma and Jimmy Ba. 2015.
\newblock Adam: A method for stochastic optimization.
\newblock \emph{ICLR}.

\bibitem[{Kudo and Richardson(2018)}]{kudo2018sentencepiece}
Taku Kudo and John Richardson. 2018.
\newblock Sentencepiece: A simple and language independent subword tokenizer
  and detokenizer for neural text processing.
\newblock \emph{arXiv preprint arXiv:1808.06226}.

\bibitem[{Lahiri(2014)}]{lahiri2014complexity}
Shibamouli Lahiri. 2014.
\newblock \href {http://www.aclweb.org/anthology/E14-3011} {{Complexity of Word
  Collocation Networks: A Preliminary Structural Analysis}}.
\newblock In \emph{Proceedings of the Student Research Workshop at the 14th
  Conference of the European Chapter of the Association for Computational
  Linguistics}, pages 96--105, Gothenburg, Sweden. Association for
  Computational Linguistics.

\bibitem[{Lample et~al.(2018{\natexlab{a}})Lample, Conneau, Denoyer, and
  Ranzato}]{lample2018unsupervised}
Guillaume Lample, Alexis Conneau, Ludovic Denoyer, and Marc'Aurelio Ranzato.
  2018{\natexlab{a}}.
\newblock \href {https://openreview.net/forum?id=rkYTTf-AZ} {Unsupervised
  machine translation using monolingual corpora only}.
\newblock In \emph{International Conference on Learning Representations}.

\bibitem[{Lample et~al.(2018{\natexlab{b}})Lample, Ott, Conneau, Denoyer, and
  Ranzato}]{lample2018phrase}
Guillaume Lample, Myle Ott, Alexis Conneau, Ludovic Denoyer, and Marc{'}Aurelio
  Ranzato. 2018{\natexlab{b}}.
\newblock \href {https://www.aclweb.org/anthology/D18-1549} {Phrase-based {\&}
  neural unsupervised machine translation}.
\newblock In \emph{Proceedings of the 2018 Conference on Empirical Methods in
  Natural Language Processing}, pages 5039--5049, Brussels, Belgium.
  Association for Computational Linguistics.

\bibitem[{Lichtarge et~al.(2019)Lichtarge, Alberti, Kumar, Shazeer, Parmar, and
  Tong}]{lichtarge2019corpora}
Jared Lichtarge, Chris Alberti, Shankar Kumar, Noam Shazeer, Niki Parmar, and
  Simon Tong. 2019.
\newblock Corpora generation for grammatical error correction.
\newblock \emph{arXiv preprint arXiv:1904.05780}.

\bibitem[{Lichtarge et~al.(2018)Lichtarge, Alberti, Kumar, Shazeer, and
  Parmar}]{lichtarge2018weakly}
Jared Lichtarge, Christopher Alberti, Shankar Kumar, Noam Shazeer, and Niki
  Parmar. 2018.
\newblock Weakly supervised grammatical error correction using iterative
  decoding.
\newblock \emph{arXiv preprint arXiv:1811.01710}.

\bibitem[{Luong et~al.(2015)Luong, Pham, and Manning}]{luong2015effective}
Thang Luong, Hieu Pham, and Christopher~D. Manning. 2015.
\newblock Effective approaches to attention-based neural machine translation.
\newblock In \emph{EMNLP}.

\bibitem[{Merity et~al.(2016)Merity, Xiong, Bradbury, and
  Socher}]{merity2016pointer}
Stephen Merity, Caiming Xiong, James Bradbury, and Richard Socher. 2016.
\newblock Pointer sentinel mixture models.
\newblock \emph{arXiv preprint arXiv:1609.07843}.

\bibitem[{Mizumoto et~al.(2011)Mizumoto, Komachi, Nagata, and
  Matsumoto}]{mizumoto2011mining}
Tomoya Mizumoto, Mamoru Komachi, Masaaki Nagata, and Yuji Matsumoto. 2011.
\newblock Mining revision log of language learning sns for automated japanese
  error correction of second language learners.
\newblock In \emph{Proceedings of 5th International Joint Conference on Natural
  Language Processing}, pages 147--155.

\bibitem[{Ng et~al.(2014)Ng, Wu, Briscoe, Hadiwinoto, Susanto, and
  Bryant}]{ng2014conll}
Hwee~Tou Ng, Siew~Mei Wu, Ted Briscoe, Christian Hadiwinoto, Raymond~Hendy
  Susanto, and Christopher Bryant. 2014.
\newblock The conll-2014 shared task on grammatical error correction.
\newblock In \emph{Proceedings of the Eighteenth Conference on Computational
  Natural Language Learning: Shared Task}, pages 1--14.

\bibitem[{Ott et~al.(2019)Ott, Edunov, Baevski, Fan, Gross, Ng, Grangier, and
  Auli}]{ott2019fairseq}
Myle Ott, Sergey Edunov, Alexei Baevski, Angela Fan, Sam Gross, Nathan Ng,
  David Grangier, and Michael Auli. 2019.
\newblock fairseq: A fast, extensible toolkit for sequence modeling.
\newblock \emph{arXiv preprint arXiv:1904.01038}.

\bibitem[{Paszke et~al.(2017)Paszke, Gross, Chintala, Chanan, Yang, DeVito,
  Lin, Desmaison, Antiga, and Lerer}]{paszke2017automatic}
Adam Paszke, Sam Gross, Soumith Chintala, Gregory Chanan, Edward Yang, Zachary
  DeVito, Zeming Lin, Alban Desmaison, Luca Antiga, and Adam Lerer. 2017.
\newblock Automatic differentiation in pytorch.
\newblock In \emph{NIPS-W}.

\bibitem[{Ruder(2019)}]{ruder2019neural}
Sebastian Ruder. 2019.
\newblock \emph{Neural Transfer Learning for Natural Language Processing}.
\newblock Ph.D. thesis, NATIONAL UNIVERSITY OF IRELAND, GALWAY.

\bibitem[{Sakaguchi et~al.(2017)Sakaguchi, Post, and
  Van~Durme}]{sakaguchi2017grammatical}
Keisuke Sakaguchi, Matt Post, and Benjamin Van~Durme. 2017.
\newblock Grammatical error correction with neural reinforcement learning.
\newblock In \emph{Proceedings of the Eighth International Joint Conference on
  Natural Language Processing (Volume 2: Short Papers)}, volume~2, pages
  366--372.

\bibitem[{See et~al.(2017)See, Liu, and Manning}]{see2017get}
Abigail See, Peter~J. Liu, and Christopher~D. Manning. 2017.
\newblock \href {https://doi.org/10.18653/v1/P17-1099} {Get to the point:
  Summarization with pointer-generator networks}.
\newblock In \emph{Proceedings of the 55th Annual Meeting of the Association
  for Computational Linguistics (Volume 1: Long Papers)}, pages 1073--1083,
  Vancouver, Canada. Association for Computational Linguistics.

\bibitem[{Sennrich et~al.(2016{\natexlab{a}})Sennrich, Haddow, and
  Birch}]{sennrich2016improving}
Rico Sennrich, Barry Haddow, and Alexandra Birch. 2016{\natexlab{a}}.
\newblock \href {https://doi.org/10.18653/v1/P16-1009} {Improving neural
  machine translation models with monolingual data}.
\newblock In \emph{Proceedings of the 54th Annual Meeting of the Association
  for Computational Linguistics (Volume 1: Long Papers)}, pages 86--96, Berlin,
  Germany. Association for Computational Linguistics.

\bibitem[{Sennrich et~al.(2016{\natexlab{b}})Sennrich, Haddow, and
  Birch}]{sennrich2016neural}
Rico Sennrich, Barry Haddow, and Alexandra Birch. 2016{\natexlab{b}}.
\newblock Neural machine translation of rare words with subword units.
\newblock In \emph{ACL}.

\bibitem[{Sutskever et~al.(2014)Sutskever, Vinyals, and
  Le}]{sutskever2014sequence}
Ilya Sutskever, Oriol Vinyals, and Quoc~V Le. 2014.
\newblock Sequence to sequence learning with neural networks.
\newblock In \emph{NIPS}.

\bibitem[{Tajiri et~al.(2012)Tajiri, Komachi, and Matsumoto}]{tajiri2012tense}
Toshikazu Tajiri, Mamoru Komachi, and Yuji Matsumoto. 2012.
\newblock Tense and aspect error correction for esl learners using global
  context.
\newblock In \emph{Proceedings of the 50th Annual Meeting of the Association
  for Computational Linguistics: Short Papers-Volume 2}, pages 198--202.
  Association for Computational Linguistics.

\bibitem[{Vaswani et~al.(2017)Vaswani, Shazeer, Parmar, Jones, Uszkoreit,
  Gomez, and Kaiser}]{vaswani2017attention}
Ashish Vaswani, Noam Shazeer, Niki Parmar, Llion Jones, Jakob Uszkoreit,
  Aidan~N Gomez, and \L~ukasz Kaiser. 2017.
\newblock Attention is all you need.
\newblock In \emph{NIPS}.

\bibitem[{Vincent et~al.(2008)Vincent, Larochelle, Bengio, and
  Manzagol}]{vincent2008extracting}
Pascal Vincent, Hugo Larochelle, Yoshua Bengio, and Pierre-Antoine Manzagol.
  2008.
\newblock Extracting and composing robust features with denoising autoencoders.
\newblock In \emph{Proceedings of the 25th international conference on Machine
  learning}, pages 1096--1103. ACM.

\bibitem[{Wang and Zheng(2015)}]{wang2015transfer}
Dong Wang and Thomas~Fang Zheng. 2015.
\newblock Transfer learning for speech and language processing.
\newblock In \emph{2015 Asia-Pacific Signal and Information Processing
  Association Annual Summit and Conference (APSIPA)}, pages 1225--1237. IEEE.

\bibitem[{Xie et~al.(2018)Xie, Genthial, Xie, Ng, and
  Jurafsky}]{xie2018noising}
Ziang Xie, Guillaume Genthial, Stanley Xie, Andrew Ng, and Dan Jurafsky. 2018.
\newblock \href {https://doi.org/10.18653/v1/N18-1057} {Noising and denoising
  natural language: Diverse backtranslation for grammar correction}.
\newblock In \emph{Proceedings of the 2018 Conference of the North American
  Chapter of the Association for Computational Linguistics: Human Language
  Technologies, Volume 1 (Long Papers)}, pages 619--628, New Orleans,
  Louisiana. Association for Computational Linguistics.

\bibitem[{Yannakoudakis et~al.(2018)Yannakoudakis, Andersen, Geranpayeh,
  Briscoe, and Nicholls}]{yannakoudakis2018developing}
Helen Yannakoudakis, {\O}istein~E Andersen, Ardeshir Geranpayeh, Ted Briscoe,
  and Diane Nicholls. 2018.
\newblock Developing an automated writing placement system for esl learners.
\newblock \emph{Applied Measurement in Education}, 31(3):251--267.

\bibitem[{Zhao et~al.(2019)Zhao, Wang, Shen, Jia, and Liu}]{zhao2019improving}
Wei Zhao, Liang Wang, Kewei Shen, Ruoyu Jia, and Jingming Liu. 2019.
\newblock Improving grammatical error correction via pre-training a
  copy-augmented architecture with unlabeled data.
\newblock In \emph{NAACL}.

\end{thebibliography}
\bibliographystyle{acl_natbib}

\appendix

\section{Copy-Augmented Transformers: Formal Derivation}\label{app:copytransformer}

Copy-augmented Transformers \cite{zhao2019improving} incorporate an attention-based copying mechanism \cite{gu2016incorporating, see2017get, jia2016data} in the decoder of Transformers. 
For each output token $y_t$ at output position $t$, given source token sequence ${\bf x} = (x_1, \dotsc, x_{T'})$, the output probability distribution over token vocabulary $V$ is defined as:
\begin{align}
    {\bf H}^\mathrm{enc} &= \mathrm{enc}({\bf x}) \\
    {\bf h}_t^\mathrm{dec} &= \mathrm{dec}\left(y_{1:t-1}; {\bf H}^\mathrm{enc}\right) \\
    p^\mathrm{gen}(y_t \mid y_{1:t-1}; {\bf x}) &= \mathrm{softmax}\left( {\bf W}^\mathrm{gen} {\bf h}_t^\mathrm{dec} \right) \label{eqn:generative}
\end{align}
where $\mathrm{enc}$ denotes the encoder that maps the source token sequence ${\bf x}$ to a sequence of hidden vectors ${\bf H}^\mathrm{enc} \in \mathbb{R}^{d \times T'}$, $\mathrm{dec}$ denotes the decoder that takes output tokens at previous time steps along with encoded embeddings and produces a hidden vector $ {\bf h}_t^\mathrm{dec} \in  \mathbb{R}^d$, and ${\bf W}^\mathrm{gen} \in \mathrm{R}^{|V| \times d}$ is a learnable linear output layer that maps the hidden vector to a pre-softmax output probabilities (``logits'').
We denote the resulting distribution as the (token) generative distribution, denoted as $p^\mathrm{gen}$.

A copy attention layer can be defined as an additional (possibly multi-head) attention layer between the encoder outputs and the final-layer hidden vector at the current decoding step.
The attention layer yields two outcomes, the layer output ${\bf o}_t$ and the corresponding attention scores ${\bf s}_t$:
\begin{align}\label{eqn:attention}
    {\bf s}_t &= \mathrm{softmax}\left( \frac{({\bf h}_t^\mathrm{dec})^T{\bf H}^\mathrm{enc}}{\sqrt{d}} \right) \\ 
    {\bf o}_t &= {\bf H}^\mathrm{enc} {\bf s}_t 
\end{align}
The copy distribution is then defined as the attention scores in \eqref{eqn:attention} themselves\footnote{In practice, this involves adding up the copy scores defined for each source token into a $|V|$-dimensional vector, using commands such as \texttt{scatter\_add()} in PyTorch.}:
\begin{equation}\label{eqn:copyattn}
    p^\mathrm{copy}(y_t \mid y_{1:t-1}; {\bf x}) = {\bf s}_t
\end{equation}

The final output of a copy-augmented Transformer as a mixture of both generative and copy distributions.
The mixture weight\footnote{When computing the mixture weight $\alpha_t^\mathrm{copy}$, \citet{zhao2019improving} applies a linear layer to ${\bf H}^\mathrm{enc}\tilde{\bf s}_t$, where $\tilde{\bf s}_t$ are the attention scores in \eqref{eqn:attention} \textit{before} taking softmax. 
Our formulation gives essentially the same copying mechanism, while being more compatible to standard Transformer implementations.} $\alpha_t^\mathrm{copy}$ is defined at each decoding step as follows:
\begin{align}
    \alpha_t^\mathrm{copy} &= \mathrm{sigmoid} \left( ({\bf w}^\mathrm{alpha})^T {\bf o}_t \right) \\
    p(y_t) &= (1 - \alpha_t^\mathrm{copy}) \cdot p^\mathrm{gen}(y_t) + \alpha_t^\mathrm{copy} \cdot p^\mathrm{copy}(y_t) \label{app:eqn:copyaugmented}
\end{align}
where ${\bf w}^\mathrm{alpha} \in \mathbb{R}^d$ is a learnable linear output layer. 
(For simplicity, we omit the dependencies of all probabilities in \eqref{app:eqn:copyaugmented} on both $y_{1:t-1}$ and ${\bf x}$.)
The mixture weight balances between how likely it is for the model to simply copy a source token, rather than generating a possibly different token.

\section{Exploratory Data Analysis}\label{app:exploratory}

\subsection{Data Sizes}

\begin{figure}[t]
    \centering
    \includegraphics[width=7.5cm]{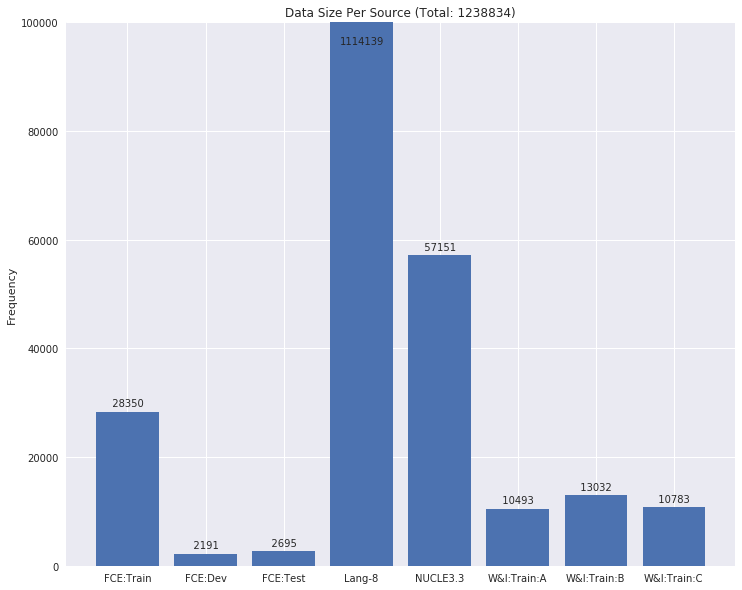}
    \caption{Data size per source for all Restricted Track training data. Number includes multiple annotations for Lang-8. Vertical axis is capped at 100K for a better visual comparison among the smaller sources. The three FCE splits (train, dev, test) are collectively used for training, and the three W\&I+L splits correspond to three English proficiency levels (``A'', ``B'', ``C''). After duplicate removal, only 575K of the Lang-8 parallel corpus are actually used for training.}
    \label{fig:data_sizes}
\end{figure}

Figure \ref{fig:data_sizes} illustrates the number of available parallel corpora (counting multiple annotations) across data sources.
Note that the vertical axis is capped at 100K for a better visual comparison among other sources.

For the Lang-8 dataset, we count all available (ranging from 1 to 8) annotations for each of 1.04M original sentences.
Also note that we only use the subset of Lang-8 whose source and target sentences are different, leaving only 575K sentences instead of 1.11M.

\subsection{Sentence Length vs. Number of Edits}

\begin{figure*}[t]
    \centering
    \includegraphics[width=16cm]{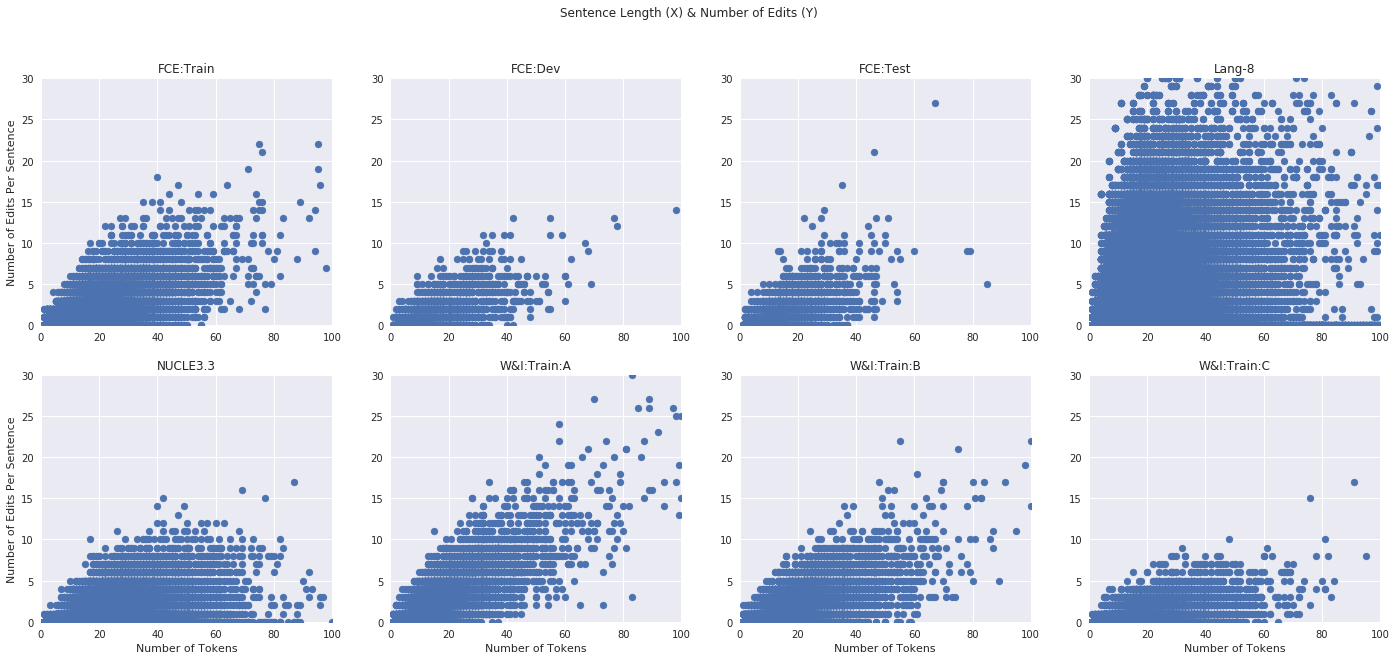}
    \caption{Sentence length versus the number of edits made in each sentence, across all training data sources for the Restricted Track. The horizontal axis is capped at 100 words (less than 0.02\% of all sentences contain more than 100 words). The vertical axis is capped at 40 edits (less than 0.02\% of all sentences contain more than 30 edits).}
    \label{fig:lengths_and_edits}
\end{figure*}

Figure \ref{fig:lengths_and_edits} illustrates the distribution of sentence lengths and the number of edits per sentence across different data sources. 

Table \ref{tbl:data_diff} includes our permutation test\footnote{We used the off-the-shelf \texttt{mlxtend} package to run permutation tests. See \url{http://rasbt.github.io/mlxtend/user_guide/evaluate/permutation_test/}.}
 results on the number of edits per sentence, normalized by sentence length (i.e., number of word-level tokens), between training data sources. 
Using an approximate permutation test with 10k simulations and a significant level of $\alpha = 0.05$, we find that there is a statistical difference in the normalized edit count per sentence between the W\&I training set and each of FCE, NUCLE, and Lang-8. 
This serves as a preliminary experiment showing how the distribution of grammatical errors can be significantly different across different sources -- even when they belong to a roughly similar domain.

\begin{table}[t]
    \centering
    \begin{tabular}{cccc}
        \Xhline{1.1pt}
        \multirowcell{2}{\bf Source} & \multirowcell{2}{\bf \# Sent.} & {\bf \# Edits} & {\bf Perm. Test} \\
         & & {\bf / Length} & {\bf vs. W\&I} \\ \hline
        W\&I-Train & 34.3K & 0.0922 & n/a \\ \hline
        FCE        & 33.2K & 0.0898 & $p = .002$ \\ 
        NUCLE      & 57.2K & 0.0318 & $p < .001$ \\ 
        Lang-8     & 1.11M & 0.1357 & $p < .001$ \\ 
        \Xhline{1.1pt}
    \end{tabular}
    \caption{Comparing the average number of edits per sentence, normalized by sentence length, between the W\&I training set and other available training data sources for the Restricted Track. ``vs. W\&I'' refers to the result of an approximate permutation test (10k rounds) against that in the W\&I training set. Under the significance level of $\alpha = 0.05$, the number for FCE, NUCLE, and Lang-8 are all significantly different from that for the W\&I training set.}
    \label{tbl:data_diff}
\end{table}

\section{Full Noising Algorithm}

Algorithms \ref{alg:constructing} and \ref{alg:noising} detail our noising scenarios.

\begin{algorithm*}
\begin{algorithmic}
\Function{ConstructNoiseDictionary}{ParallelCorpus, min\_count}
\State Initialize a dictionary dict
\For{(CorToken, OriToken) in ParallelCorpus}
    \State dict[CorToken] += OriToken
\EndFor

\For{CorToken, OriTokenList in dict}
    \For{OriToken in OriTokenList}
        \If{count(OriToken) $<$ min\_count}
            \State delete OriToken from dict[CorToken]
        \EndIf
    \EndFor
    \If{length(OriTokenList)==1 and CorToken==OriTokenList[0]}
        \State delete OriToken from dict
    \EndIf
\EndFor

\Return dict
\EndFunction
\caption{Pseudocode for constructing noise dictionary}
\label{alg:constructing}
\end{algorithmic}
\end{algorithm*}

\begin{algorithm*}
\begin{algorithmic}
\Function{change\_type}{word, prob}
    \State preposition\_set = [$\emptyset$, for, to, at, $\cdots$]
    \If{random[0, 1] $>$ prob}
        \Return word
    \Else
        \If{word in preposition\_set}
            \State random\_choose\_one\_from(preposition\_set)
        \ElsIf{word is Noun} change\_number(word)
        \ElsIf{word is Verb} change\_form(word)
        \EndIf
    \EndIf
    
    \Return word
\EndFunction

\Function{make\_noise}{sentence, prob}
    \State dict = ConstructNoiseDictionary(ParallelCorpus, min\_count)
    \State noised = []
    \For{word in sentence}
        \If{word in dict and random[0, 1] $>$ prob}
            \State candidates = dict[word]
            \State noise = random\_choose\_one\_from(candidates)
        \Else
            \State noise = change\_type(word)
        \EndIf
        \State noised += noise
    \EndFor
    \Return noised
\EndFunction

\caption{Pseudocode for generating noisy sentences}
\label{alg:noising}
\end{algorithmic}
\end{algorithm*}

\section{Results on error categories}
Table \ref{tbl:error_categories} shows the result on error categories.
\begin{table}
    \centering
    \begin{tabular}{cccc}
        \Xhline{1.1pt}
        \textbf{Error types}  & \textbf{P} & \textbf{R} & \textbf{F0.5} \\
        \hline
         ADJ & 71.43 & 28.57 & 54.95 \\
         ADJ:FORM & 100.00 & 40.00 & 76.92 \\
         ADV & 70.59 &	22.22 &	49.18 \\
         CONJ & 100.00 &	4.76 &	20.00 \\
         CONTR & 100.00 &	91.67 &	98.21 \\
         DET & 78.95 &	47.04 &	69.52 \\
         MORPH & 81.18 &	49.29 &	71.88 \\
         NOUN & 64.52 &	17.24 &	41.67 \\
         NOUN:INFL & 100.00 &	41.18 &	77.78 \\
         NOUN:POSS & 81.82 &	48.21 &	71.81 \\
         ORTH & 87.38 &	49.60 &	75.83 \\
         OTHER & 55.93 &	15.30 &	36.53 \\
         PART & 76.19 &	55.17 &	70.80 \\
         PREP & 69.69 &	49.01 &	64.27 \\
         PRON & 78.67 &	43.70 &	67.82 \\
         PUNCT & 79.95 &	70.48 &	77.86 \\
         SPELL & 76.07 &	82.41 &	77.26 \\
         VERB & 66.67 &	7.88 &	26.76 \\
         VERB:FORM & 72.45 &	73.20 &	72.60 \\
         VERB:INFL & 100.00 &	85.71 &	96.77 \\
         VERB:SVA & 83.77 &	85.43 &	84.09 \\
         VERB:TENSE & 71.43 &	45.64 &	64.18 \\
         WO & 67.74 &	25.61 &	50.97 \\
         \Xhline{1.1pt}
    \end{tabular}
    \caption{Results on error types. }
    \label{tbl:error_categories}
\end{table}

\section{CoNLL-2014 Full Results \\ (Without Using W\&I+L)}\label{app:comparison}

In Table \ref{tbl:track0training}, we include the training progress for our result for CoNLL-2014.

A noticeable difference between this result and our results for Restricted Track and Low Resource Track is that adaptation via fine-tuning is not necessarily effective here.
We hypothesize that this is mostly due to the fact that the training subset to which we fine-tune our model (NUCLE) comes from a different source than the actual test set (CoNLL-2014) -- despite the fact that both datasets have similar domains (personal essays from English students), they can still have many other different characteristics, including the writer's English proficiency and annotation styles.

\begin{table}[t]
    \centering
    \begin{tabular}{lcc>{\bfseries}c}
        \Xhline{1.1pt}
        \multirowcell{2}{\textbf{Steps}} & \multicolumn{3}{c}{\textbf{CoNLL-2014}} \\
         & \textbf{P} & \textbf{R} & \textbf{F0.5}  \\ \hline
        Spellcheck         & 54.75 &  5.75 & 20.25   \\ \hline
        + Pre-train (b)    & 54.76 & 15.09 & 35.89  \\ 
        + Train (b)        & 60.43 & 34.22 & 52.40  \\ 
        + Fine-tune (b)    & 60.81 & 33.32 & 52.20  \\ \hline
        + Pre-train (c)    & 65.81 & 24.17 & 48.95  \\ 
        + Train (c)        & 61.38 & 30.97 & 51.30  \\ 
        + Fine-tune (c)    & 60.82 & 32.50 & 51.79  \\ \hline
        + Ensemble (b+c)   & 71.11 & 32.56 & 57.50  \\ 
        \Xhline{1.1pt}
    \end{tabular} 
    \caption{Training progress on CoNLL-2014. \textit{No W\&I+Locness datasets were used in these results.} `b' and `c' refer to the base and copy configurations of the Transformer, respectively. Evaluation is done using the MaxMatch ($M^2$) scorer. Pre-processing \& post-processing are included before the first step and after the last step, respectively.}
    \label{tbl:track0training}
\end{table}

\section{Training Details}
\label{app:training_detail}
Our model training is a three-stage process: DAE pre-training, training, and fine-tuning, except in Low Resource Track where there is no fine-tuning data. 
At each step, we train a model until its ERRANT score on the development set reaches convergence, and use the learned weights as initial values for the next step. 
For pre-training, we used a learning rate of $5 \cdot 10^{-4}$ for the base and copy-augmented Transformers and $10^{-3}$ for the large Transformer.
For training, we reset the optimizer and set the learning rate to $10^{-4}$. For fine-tuning (if available), we again reset the optimizer and set the learning rate to $5 \cdot 10^{-5}$. 
In all training steps, we used the Adam \cite{kingma2014adam} optimizer with the inverse square-root schedule and a warmup learning rate of $10^{-7}$, along with a dropout rate of $0.3$.

\section{Further Analysis}

\subsection{Effect of Copying Mechanisms \& Ensembles}\label{sec:effect_copy}

\begin{table}[b]
    \centering
    \begin{tabular}{cccc}
        \Xhline{1.1pt}
        \multirowcell{2}{\bf Model (Config.)} & \multicolumn{3}{c}{\textbf{W\&I+L Test}} \\
         & \textbf{P} & \textbf{R} & \textbf{F0.5} \\ \hline
        Vanilla (Large) & 63.66 & 56.82 & 62.17  \\
        Copy (Copy)     & 66.02 & 53.41 & 63.05  \\ \hline
        $\Delta$        & +2.36 & -3.41 & +0.88  \\
        \Xhline{1.1pt}
    \end{tabular} 
    \caption{Single-model ERRANT scores for Restricted Track, using a large Transformer and a copy-augmented Transformer.} 
    \label{tbl:vanilla_vs_copy}
\end{table}

One of our contributions is to highlight the benefit of ensembling multiple models with diverse characteristics.
As shown in Table 3, the final ensemble step involving different types of models was crucial for our model's performance, improving the test score by over 6 $F_{0.5}$ points. 
We first noticed that the copy-augmented Transformer learns to be more conservative -- i.e., higher precision but lower recall given similar overall scores -- in its edits than the vanilla Transformer, presumably because the model includes an inductive bias that favors copying (i.e., not editing) the input token via its copy attention scores.
Table \ref{tbl:vanilla_vs_copy} shows this phenomenon for Restricted Track.

Given multiple models with diverse characteristics, the choice of models for ensemble can translate to controlling how conservative we want our final model to be.
For example, combining one vanilla model with multiple independent copy-augmented models will result in a more conservative model.
This could serve as an alternative to other methods that control the precision-recall ratio, such as the edit-weighted loss \cite{junczys2018approaching}.

\end{document}